\documentclass[10pt, conference]{IEEEtran}
\IEEEoverridecommandlockouts
\usepackage{cite}
\usepackage{amsmath,amssymb,amsfonts}
\usepackage{algorithm}
\usepackage{algpseudocode}
\usepackage{nicefrac}
\usepackage{graphicx}
\usepackage{textcomp}
\usepackage{xcolor}
\usepackage{xspace}
\usepackage{amsthm}
\usepackage{xurl}
\usepackage{thm-restate}
\usepackage{comment}
\usepackage{subfig}
\allowdisplaybreaks

\def\BibTeX{{\rm B\kern-.05em{\sc i\kern-.025em b}\kern-.08em
    T\kern-.1667em\lower.7ex\hbox{E}\kern-.125emX}}
    
\newcommand{\CBayes}{\mathcal{C}_{\mathrm{Bayes}}}
\newcommand{\gbar}{\overline{\mathbf{g}}}
\newcommand{\gtilde}{\tilde{\mathbf{g}}}

\newcommand{\VM}{\mathcal{V}}
\newcommand{\Gauss}{\mathcal{N}}
\newcommand{\Alg}[1]{Alg. \ref{#1}}
\newcommand{\Eqn}[1]{Eqn (\ref{#1})}

\newcommand{\Lem}[1]	{Lem \ref{#1}}
\newcommand{\Sec}[1]	{Sec. \ref{#1}}
\newcommand{\App}[1]	{Appendix \ref{#1}}
\newcommand{\Fig}[1]	{Figure \ref{#1}}

\newcommand{\Reals}{\mathbb{R}_{\geq 0}}
\newcommand{\calw}  {\ensuremath{\mathcal{W}}}
\newcommand{\calx}  {\ensuremath{\mathcal{X}}}
\newcommand{\caly}  {\ensuremath{\mathcal{Y}}}
\newcommand{\calz}  {\ensuremath{\mathcal{Z}}}
\newcommand{\Dist}  {\ensuremath{\mathbb{D}}}

\newcommand{\VPrior} {\ensuremath{V_g}}
\newcommand{\VPosterior} {\ensuremath{V_g}}
\newcommand{\Leak} 	{\mathcal{L}_g}
\definecolor{greenmunsell}{rgb}{0.0, 0.66, 0.47}

\theoremstyle{definition}
\newtheorem{definition}{Definition}[section]
\theoremstyle{remark}
\newtheorem{remark}{Remark}
\newtheorem{theorem}{Theorem}[section]

\newtheorem{lemma}[theorem]{Lemma}

\begin{document}

\title{Bayes' capacity as a measure for reconstruction attacks in federated learning}

\author{
\IEEEauthorblockN{Sayan Biswas\IEEEauthorrefmark{1}, Mark Dras\IEEEauthorrefmark{2}, Pedro Faustini\IEEEauthorrefmark{2}, 
Natasha Fernandes\IEEEauthorrefmark{2}, \\ Annabelle McIver\IEEEauthorrefmark{2}, Catuscia Palamidessi\IEEEauthorrefmark{3} and
Parastoo Sadeghi\IEEEauthorrefmark{4}
}
\IEEEauthorblockA{
\IEEEauthorrefmark{1}EPFL,
Lausanne, Switzerland}
\IEEEauthorblockA{
\IEEEauthorrefmark{2}Macquarie University,
Sydney, Australia}
\IEEEauthorblockA{
\IEEEauthorrefmark{3}INRIA and \'{E}cole Polytechnique,
Palaiseau, France}
\IEEEauthorblockA{
\IEEEauthorrefmark{4}UNSW,
Canberra, Australia}}

%\and

\maketitle
%\IEEEpeerreviewmaketitle

\begin{abstract}
Within the machine learning community, reconstruction attacks are a principal attack of concern and have been identified even in federated learning, which was designed with privacy preservation in mind. In federated learning, it has been shown that an adversary with knowledge of the machine learning architecture is able to infer the exact value of a training element given an observation of the weight updates performed during stochastic gradient descent. In response to these threats, the privacy community recommends the use of differential privacy in the stochastic gradient descent algorithm, termed DP-SGD. However, DP has not yet been formally established as an effective countermeasure against reconstruction attacks. In this paper, we formalise the reconstruction threat model using the information-theoretic framework of quantitative information flow. We show that the Bayes' capacity, related to the Sibson mutual information of order infinity, represents a tight upper bound on the leakage of the DP-SGD algorithm to an adversary interested in performing a reconstruction attack. We provide empirical results demonstrating the effectiveness of this measure for comparing mechanisms against reconstruction threats.
\end{abstract}


\section{Introduction}
\label{s1827}

Reconstruction attacks - in which an attacker can reconstruct the training data used to build a machine learning model - are a principal attack of concern in machine learning, since training data are often considered to be sensitive. These attacks are of particular concern in Federated Learning (FL), a distributed form of machine learning in which weaker machine learners collaborate to train a more powerful model without having to share their training data. 

Numerous works~\cite{zhu-etal:2019:NeurIPS, yin-etal:2021:CVPR, geiping-etal:2020:NeurIPS} have shown that FL is susceptible to %particular attacks %This has necessitated the use of privacy-preserving methods in the vulnerable stochastic gradient descent algorithm in FL. Differential privacy (DP) is typically the privacy-preserving technique of choice, leading to the well-known DP-SGD algorithm (\Alg{alg:sgd}) which is now standard in privacy-preserving machine learning.
%
%These are called 
\emph{gradient-based} reconstruction attacks. These describe an attacker who can observe the gradients produced in the stochastic gradient descent algorithm, and by incorporating their knowledge of the architecture, can reconstruct the sensitive input data. Significant works in this regard are \emph{deep leakage from gradients} by Zhu et al.~\cite{zhu-etal:2019:NeurIPS}, \emph{see through gradients} by Yin et al.~\cite{yin-etal:2021:CVPR} and \emph{inverting gradients} by Geiping et al.~\cite{geiping-etal:2020:NeurIPS}. Such attacks are feasible in distributed learning architectures such as FL because the gradients computed by each client are shared with a central server who, in the threat model we consider, is an honest-but-curious adversary.

We observe that theoretical work in this area~\cite{zhu2021r,gong2023gradient} differs from practical work in understanding \emph{reconstruction}, particularly when the sensitive input data are images. Theoretically, a reconstruction attack occurs when an input data value is inferred \emph{exactly}; practically, a reconstruction attack occurs when a sensitive image can be reconstructed \emph{approximately}. This has implications for the information-theoretic measures used to describe leakage, which we discuss later in the paper.

A second important caveat is that the reconstruction attacks described in the literature assume that no privacy protections have been put in place. In this paper, we study reconstruction attacks with differential privacy applied to the gradients as in \Alg{alg:sgd}, also known as DP-SGD. %(Note: In this algorithm the {\color{blue} blue} components are specific to DP-SGD, the black components form the usual SGD algorithm.) 

While it is standard practice in the DP literature to compare mechanisms by ``comparing their epsilons", we argue that a formal information-theoretic analysis of reconstruction attacks reveals that the Bayes' capacity better aligns with the attacker's strategy. The Bayes' capacity is known in the quantitative information flow literature as a robust upper bound on the leakage of a system against any Bayesian attacker~\cite{Alvim20:Book}; its logarithm is known in the information-theory literature as the Sibson mutual information of order $\alpha = \infty$~\cite{sibson1969information}. We make the following contributions:
\begin{enumerate}
\item We formalise reconstruction attacks and show that Bayes' capacity acts as a measure of attack success. 
\item We introduce the continuous version of the Bayes' capacity and derive a formulation for the Bayes' capacity of the DP-SGD algorithm.
\item We propose an alternative mechanism for DP-SGD based on the Von Mises-Fisher distribution and compute its Bayes' capacity in the DP-SGD algorithm.
\item We experimentally compare the Bayes' capacity and the $\epsilon$ values for the above mechanisms against proxies for reconstruction attacks, showing that the attacks correlate well with Bayes' capacity compared with $\epsilon$.
\end{enumerate}

Proofs for our results can be found in \App{app:proofs}.

%\section{Reconstruction Attacks in Federated Learning}
%In this section we provide an analysis of reconstruction attacks in the literature and propose an alternative measure which corresponds with an adversarial scenario in which an attacker wishes to reconstruct the training datum given an observation from the machine learning system.

\section{A Formal Model for Reconstruction Attacks}\label{sec:measuring}

% Perfect reconstruction vs reconstruction. Focus on strengths of Bayes capacity and what it measures.

In this section, we use a theoretical understanding of reconstruction - inferring the \emph{exact} value of the secret - to formalise a leakage model using on the information-theoretic framework of quantitative information flow~\cite{Alvim20:Book}.
%We begin with a formal model for the attacks described in the literature, which we then extend to the DP-SGD algorithm.

\begin{algorithm}[!t]
\caption{DP-SGD with Gaussian noise}\label{alg:sgd}
\begin{algorithmic}[1]
\State \textbf{Input:} Examples $\{x_1,\ldots,x_N\}$, loss function $\mathcal{L}(\theta) = \frac{1}{N} \sum_i \mathcal{L}(\theta, x_i)$. Parameters: learning rate $\eta_t$, noise scale $\sigma$, group size $L$, gradient norm bound $c$.
\State \textbf{Initialise} $\theta_0$ randomly
\For{$t \in T$}
   \State $L_t \gets $ random sample of $L$ indices from $1{\ldots}N$
   \For{$i \in L_t$}
   	\State $\mathbf{g}_t(x_i) \gets  \nabla_{\theta_t} \mathcal{L}(\theta_t, x_i)$
    \Comment{Compute gradient}
	\State $\gbar_t(x_i) \gets \nicefrac{\mathbf{g}_t(x_i)}{\max (1, \frac{\| \mathbf{g}_t(x_i)\|_2}{c}) }$
\Comment{Clip gradient}
  \EndFor  
   \State $\gtilde_t \gets \frac{1}{L} \sum_i \gbar_t(x_i)$
   \Comment{Average}
   \State $\gtilde_t \gets \gtilde_t + \frac{1}{L} \Gauss(0, c^2\sigma^2)$
   \Comment{Add noise}
   {\color{red}\State // Leak $\gtilde_t$}
   \State $\theta_{t+1} \gets \theta_t - \eta_t \gtilde_t$
    \Comment{Descent}
\EndFor
\State \textbf{Output} $\theta_T$ %and compute the overall privacy cost.
\end{algorithmic}
\end{algorithm}

\subsection{Model for the machine learning system}\label{sec:model}

%The stochastic gradient descent algorithm is described in \Alg{alg:sgd}. In the FL attacks we consider, the examples $\{x_1,\ldots,x_N\}$ are distributed across $N$ clients; a central server computes the initial random weight assignment (line 2) and sends it to each client; each selected client (indices in $L_t$) computes a gradient for its example (line 6), and these gradients are sent back to the server to be averaged (line 10). The (untrusted) central server performs a reconstruction attack by observing a gradient $\mathbf{g_t}$ and inferring the training element $x_i$ that generated it. 

The standard DP-SGD algorithm is depicted in \Alg{alg:sgd}. In federated learning, DP-SGD is performed in a distributed fashion: each client begins with their own training examples (inputs $x_1,...,x_N$) and a common loss function ($\mathcal{L}$); the server performs a weight initialisation (line 2) which is shared with the clients, who each perform the gradient update steps over their training batch (lines 4-10) and then send back the gradients to the server for the gradient descent step (line 12). The updates ($\theta_{t+1}$) are shared with the clients for the next ($t+1$) round. The leak statement (line 11) indicates the information observed by the adversary, who in this scenario is the server. %We remark that the standard SGD algorithm (without differential privacy) excludes lines 7 (clipping) and 10 (noise addition).

%We use information-theoretic framework of quantitative information flow to model the information flow to the attacker through the system. 
Formally, we model a system as an information-theoretic channel $C:\calx \rightarrow \Dist{\caly}$, taking secrets $\calx$ to distributions over observations $\caly$ can be written as a matrix $C:{\calx \times \caly}\rightarrow [0,1]$ whose rows are labelled with secrets, and columns labelled with observations, and where $C_{x, y}$ is the probability of observing $y \in \caly$ given the secret $x \in \calx$. 

Applying this model to \Alg{alg:sgd}, we have that each client can be expressed as a channel taking a set of $L$ inputs $x_L \in \calx^L$ (line 4) and producing an observable (noisy) averaging-of-gradients $y \in \caly$ (line 10), where we denote by $\caly$ the set of all possible observations and $\calx^L$ the set of all possible input sets of length $L$. For now, our model assumes that $\calx^L$ and $\caly$ are discrete; we extend this idea to continuous domains in \Sec{sec:bayes}. We remark that once the set of size $L$ is selected, the algorithm consists of deterministic steps (lines 5-9), followed by a probabilistic post-processing (line 10). We can model this system as a composition of channels $D \circ C$, where $C$ is the deterministic channel described by lines 5-9 and $D$ is the noise-adding mechanism from line 10. Theoretical analyses have provided conditions under which $C$ describes a 1-1 function and could therefore be inverted by an adversary~\cite{zhu2021r, gong2023gradient}. We use this as our basic assumption since this represents the maximum leakage of the system to an adversary.~\footnote{Note: If it is not the case that the model $M$ leaks the secrets exactly, then the system will be more protected, and we leave the study of this to future work.} We remark that in practice, counter-measures such as adjusting the loss function to favour more ``realistic'' images are applied to increase the reconstruction accuracy~\cite{yin-etal:2021:CVPR}, however, even in these cases approximate measures (eg. mean-squared error) are required in order to measure the effectiveness of the reconstruction attack.

We write the type of $D$ as $\caly \rightarrow \Dist{\caly}$, that is, taking gradient vectors to distributions over gradient vectors, and the type of $C$ as $\calx \rightarrow \Dist{\caly}$. Note that $\Dist{\caly}$ incorporates deterministic channels as point distributions on $\caly$.  Mathematically, the composition $D \circ C$ corresponds to $C {\cdot} D$ where $\cdot$ is matrix multiplication. This composition describes the information flow to the attacker, from secrets $\calx$ to noisy gradients ${\caly}$. 

%Note that at this point, we have not determined the structure of $D$. We will compute this later when we consider different noise-adding mechanisms which could be used in DP-SGD.  

\subsection{Model for the attacker}\label{sec:attacker}

%We use quantitative information flow to describe our attacker and introduce the basic concepts below.

We model a Bayesian attacker: they are equipped with a prior over secret inputs, modelled as a probability distribution $\pi: \Dist{\calx}$. We equip our attacker with a gain function $g:{\calw \times \calx} \rightarrow \Reals$ which describes the gain to the attacker when taking action $w \in \calw$ if the real secret is $x \in \calx$. Such gain functions can model a wide range of attacks including an attacker who wants to guess the secret exactly, who wants to guess the secret in $k$ tries or who wants to guess a value close to the secret~\cite{Alvim20:Book}. The leakage of a channel $M$ with respect to this attacker can be modelled as the difference between the attacker's expected prior and posterior knowledge, computed using their prior knowledge and their gain function. The adversary's prior knowledge of the secrets is modelled as their maximum expected gain before access to the channel $M$, given by
%
%\begin{equation}\label{eqn:prior}
$
  \VPrior(\pi) = \max_{w \in \calw} \sum_{x \in \calx} \pi_x g(w, x)  ~.
$
%\end{equation}
%
The adversary's posterior knowledge is their maximum expected gain using their knowledge of the channel $M$, derived from the Bayes rule and formulated as 
%
%\begin{equation}\label{eqn:posterior}
$
  \VPosterior(\pi, M) = \sum_{y \in \caly} \max_{w \in \calw} \sum_{x \in \calx} \pi_x M_{x, y} g(w, x)~.
$
%\end{equation}
%
The (multiplicative) leakage of the system $M$ can then be formulated as
%
%\begin{equation}\label{eqn:leakage}
$
   \Leak(\pi, M) = \frac{\VPosterior(\pi, M)}{\VPrior(\pi)}~.
$
%\end{equation}

The above models are based on standard decision-theoretic principles; further discussion can be found in~\cite{Alvim20:Book}.

\begin{remark}
We note that the above assumes discrete sets $\calx, \caly$, and indeed this assumption underlies much of the modelling for Bayesian information flows described in \cite{Alvim20:Book}. In later sections, we will extend this reasoning to continuous sets.
\end{remark}

\subsection{Model for the reconstruction attack}

In our reconstruction attack model for FL, the adversary learns the secret (training input) \emph{exactly}, using a single observation of the gradient updates. In our attacker model, this corresponds to the following gain function describing an attacker who learns the secret exactly in 1 try:

\begin{equation}\label{eqn:gain}
	g(w, x) ~=~ \left\{
		\begin{array}{ll}
			1 & \text{ if $x = w$,} \\
			0 & \text{ otherwise.}
		\end{array}
	\right.
\end{equation}

Our attacker has no knowledge of the secrets apart from that which allows them to eliminate unlikely secrets from the domain. Thus we model the attacker's prior knowledge as uniform over the (full support) domain $\calx^L$; we denote the uniform prior by $\upsilon: \Dist{\calx^L}$. Hence our reconstruction attacker's prior knowledge can be computed as
$
     \VPrior(\upsilon) = \max_{x \in \calx^L} \upsilon_x = \frac{1}{|\calx^L|}~.
$
Our attacker is assumed to have knowledge of the machine learning architecture and the DP-SGD algorithm described in \Alg{alg:sgd}. As explained above, this system can be modelled as the composition $C \cdot D$. Thus,  our attacker's posterior knowledge can be computed as
$
    \VPosterior(\upsilon, C{\cdot}D) =  \frac{1}{|\calx^L|} \sum_{z \in \calz} \max_{x \in \calx^L} (C{\cdot}D)_{x, z}.
$
%
%\noindent The last line follows from the fact that $M$ is a permutation of the identity matrix.
The leakage of the secrets via the channel $C{\cdot}D$ to the adversary is then
$
   \Leak(\upsilon, C{\cdot}D) =  \sum_{z \in \calz} \max_{x \in \calx^L} (C{\cdot}D)_{x, z}.
$
Finally, the following lemma allows a simplification of the leakage calculation for DP-SGD.

\begin{lemma}\label{lem:det}
Let $C$, $D$ be channels such that $C{\cdot}D$ is defined, and $C$ is deterministic. Then $\CBayes(C{\cdot}D) = \CBayes(D)$.
\end{lemma}

\noindent Noting that $C$ is deterministic, it follows that
\begin{equation}\label{eqn:leakage_final} 
  \Leak(\upsilon, C{\cdot}D) ~=~ \sum_{y \in \caly} \max_{x \in \caly} D_{x, y}
\end{equation}   
\noindent is a measure of the reconstruction risk for DP-SGD.

\section{Bayes' Capacity}\label{sec:bayes}

The quantity computed in \Eqn{eqn:leakage_final} is a remarkably robust measure known as the \emph{Bayes' capacity}~\cite{Alvim20:Book}. For a channel $M: \calx \rightarrow \Dist{\caly}$ it is defined:

\begin{equation}\label{eqn:bayes_capacity}
	\CBayes(M) ~=~ \sum_{y \in \caly} \max_{x \in \calx} M_{x, y}~.
\end{equation}

This quantity was recently shown to measure the maximum leakage of adversaries wanting to guess arbitrary values of the secret $\calx$~\cite{issa2019operational}. In the quantitative information flow literature~\cite{Alvim20:Book}, the Bayes' capacity is a measure of the maximum multiplicative leakage of a system to \emph{any} adversary modelled using a prior and a gain function, as described in \Sec{sec:attacker}. In the case of reconstruction attacks, this means that even if our adversarial assumptions are incorrect (eg. the adversary's prior is not uniform, or their reconstruction attack goal is to find a value close to the secret rather than the exact value of the secret), then the Bayes' capacity represents a tight upper bound on the leakage of the system to this adversary. For this reason, we consider the Bayes' capacity to be a robust and reliable measure of the effectiveness of a system to protect against reconstruction attacks.

\begin{remark}
At this point, we note that the experimental machine learning literature on image reconstruction describes reconstruction attacks as producing a ``similar'' image to the original training image, presumably because the image contains information of a sensitive nature that is inferrable from a close enough reproduction. There is no agreed-upon measure to identify similar images, so proxies such as mean-squared error (MSE)~\cite{sun-etal:2023:NeurIPS} or Structural Similarity Index (SSIM)~\cite{1284395} are used. 
However, a notable strength of Bayes' capacity lies in its comprehensive nature, as it considers all of these measures. It quantifies the maximum leakage of the system to \emph{any} adversary, regardless of the gain/loss function chosen. Thus, if Bayes' capacity is low, then any attack assessed with SSIM, MSE, or any other measure will be less successful.
%Thus, if Bayes' capacity is small, then any attack measured with SSIM or MSE (or any other measure) will be less successful.
\end{remark}

%\subsection{Bayes' Capacity as Reconstruction Attack Measure}

\noindent We summarise our findings from \Sec{sec:measuring} as follows:

\begin{enumerate}
\item The Bayes capacity $\CBayes(C{\cdot}D)$ of the machine learning system with DP-SGD $C{\cdot}D$ represents the leakage of the system to a ``reconstruction attacker'' with knowledge of the model architecture who makes a single observation of weight updates generated from the DP-SGD algorithm. When $C$ is deterministic then this is $\CBayes(D)$. 
\item Given two noise-adding mechanisms $D, D'$ applied to DP-SGD, we say that $D$ is safer than $D'$ against a reconstruction attack iff $\CBayes(C{\cdot}D) < \CBayes(C{\cdot}D')$. When $C$ is deterministic, this holds iff $\CBayes(D) < \CBayes(D')$.
\end{enumerate}

%\noindent Note that $D, D'$ could use the same noise distribution (eg. Gaussian), parametrised by different values for $\epsilon, \delta$.

\subsection{Bayes' Capacity for Continuous Mechanisms}

For the Gaussian mechanism described in \Alg{alg:sgd}, the output domain of interest is continuous, however, Bayes' capacity has previously only been defined for mechanisms acting on discrete domains. We now introduce a natural generalisation of Bayes' capacity for continuous domains.

\begin{definition}[Continuous Bayes' capacity]
Let $f: \mathcal{X} \rightarrow \Dist{\mathcal{Y}}$ be a randomised function taking inputs $x \in \mathcal{X}$ to distributions on outputs $\mathcal{Y}$. Then the Bayes' capacity of $f$ is defined as 
\begin{equation}\label{eqn:bcapacity}
    \CBayes(f) ~=~ \int_{\mathcal{Y}} \sup_x f(x)(y)~ dy
\end{equation}
where $f(x)(y)$ denotes the (continuous) probability density $f(x)$ evaluated at $y$. This is well-defined when $f$ is measurable (since the pointwise supremum of measurable functions is measurable).
\end{definition}

%We can now derive the Bayes' capacity for the DP-SGD algorithm using Gaussian noise by computing the leakage caused by lines 7-8 (aka the channel $D$ in \Eqn{eqn:leakage_final}). From \Lem{lem:det}, since line 7 is a deterministic step, we simply calculate the Bayes' capacity for the operation defined in line 8. 
We can now derive the Bayes' capacity for the DP-SGD algorithm using Gaussian noise, described by lines 4-10 of \Alg{alg:sgd}. From \Lem{lem:det}, since lines 4-9 are deterministic, we calculate the Bayes' capacity for the operation defined in line 10. We set clipping length $c=1$ as per our experiments.

\begin{theorem}[Bayes' Capacity for Gaussian]
    Let $G_{p, \sigma}: \mathcal{X} \rightarrow \Dist{\mathbb{R}^{p}}$ be the mechanism which takes as input a $p$-dimensional vector $x \in \mathcal{X}$ and outputs a perturbed vector $y \in \mathbb{R}^{p}$ by applying Gaussian noise with parameter $\sigma$ to each element of the (clipped) input vector with clipping length $c=1$, and then averaging by $L$. Then the Bayes' capacity of $G_{p, \sigma}$ is given by:
\[
    \CBayes(G_{p, \sigma}) ~=~ \frac{2}{\Gamma\left(\frac{p}{2}\right) 2^{\frac{p}{2}} \sigma^p} Z + \frac{R^p}{\Gamma\left(\frac{p}{2} + 1\right) 2^{\frac{p}{2}} \sigma^{p}}
\]
where $R$ is the number of layers in the network (each clipped to length 1) and $Z = \sum_{i=0}^{p-1} \Gamma(\frac{p-i}{2}) (\sqrt{2} \sigma)^{p-i} { p-1 \choose i } R^i$.
%\begin{proof}
%See \App{app:proofs}.
%\end{proof}
\end{theorem}

\section{Comparing Mechanisms}

\begin{algorithm}[!th]
\caption{DP-SGD with von Mises-Fisher noise}\label{alg:sgd2}
\begin{algorithmic}[1]
\State \textbf{Input:} Examples $\{x_1,\ldots,x_N\}$, loss function $\mathcal{L}(\theta) = \frac{1}{N} \sum_i \mathcal{L}(\theta, x_i)$. Parameters: learning rate $\eta_t$, noise scale $\sigma$, group size $L$, gradient norm bound $c$.
 \State \textbf{Initialise} $\theta_0$ randomly
 \For{$t \in T$}
    \State $L_t \gets $ random sample of $L$ indices from $1{\ldots}N$
    \For{$i \in L_t$}
    	\State $\mathbf{g}_t(x_i) \gets  \nabla_{\theta_t} \mathcal{L}(\theta_t, x_i)$
     \Comment{Compute gradient}
 	\State $\gbar_t(x_i) \gets \nicefrac{\mathbf{g}_t(x_i)}{\max (1, \frac{\| \mathbf{g}_t(x_i)\|_2}{c}) }$
  \Comment{Clip gradient}
   \EndFor  
  \State $\gtilde_t \gets \frac{1}{L} \sum_i \gbar_t(x_i)$  
   \Comment{Average}
   {\color{blue}  \State $\gtilde_t \gets \nicefrac{\gtilde_t}{ | \gtilde_t |}$ }
    \Comment{Scale}
    {\color{blue} \State $\gtilde_t \gets \VM(\sigma, \gtilde_t)$}
    \Comment{Add noise}
   {\color{red}\State // Leak $\gtilde_t$}
    \State $\theta_{t+1} \gets \theta_t - \eta_t \gtilde_t$
    \Comment{Descent}
 \EndFor
 \State \textbf{Output} $\theta_T$ %and compute the overall privacy cost.
\end{algorithmic}
\end{algorithm}

\begin{figure*}[th]
    \centering
    \begin{tabular}{cc}              
       % \subfloat[Epsilon vs MSE for MNIST]{\includegraphics[width=0.3\textwidth]{MNIST16_vmf_v_gauss_MSE.pdf}} \\        
        \subfloat[Epsilon vs MSE]{\includegraphics[width=0.45\textwidth]{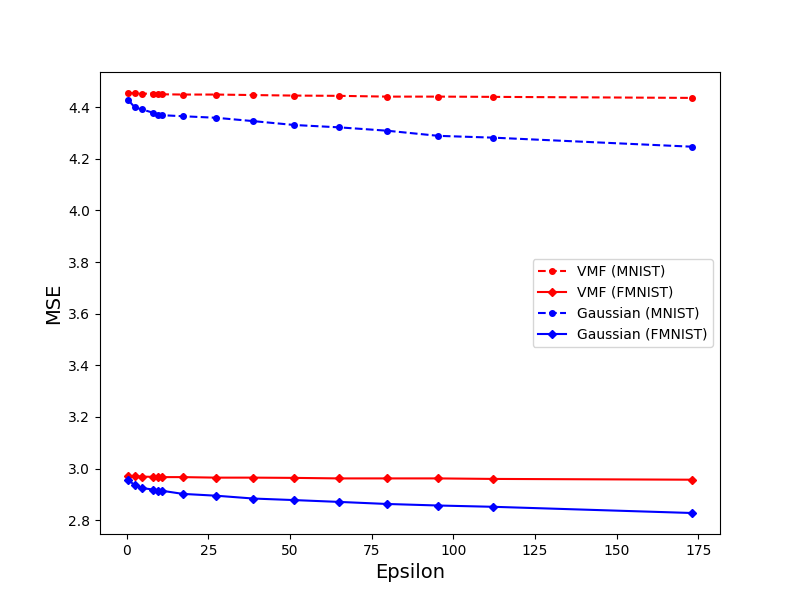}}  
        &
       % \\            
        \subfloat[Bayes' capacity vs MSE]{\includegraphics[width=0.45\textwidth]{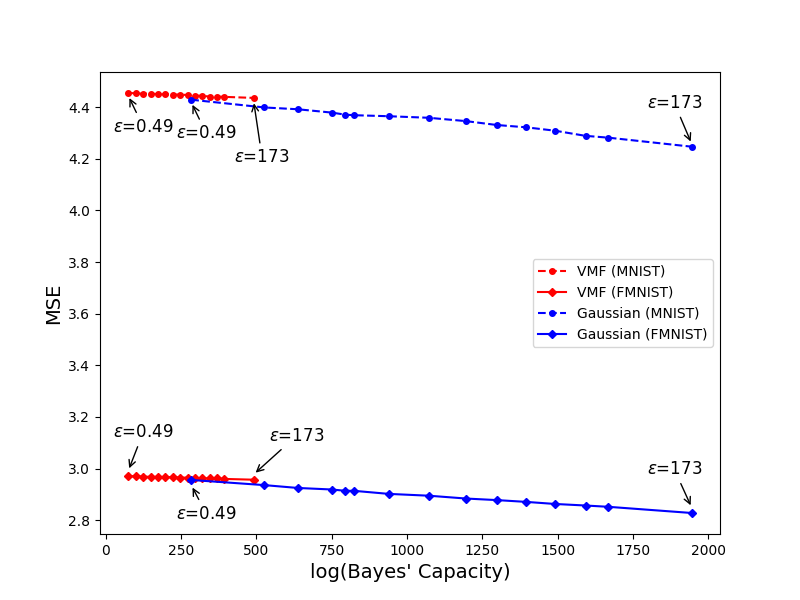}}          
\end{tabular}
   % \vspace{4mm}
    \caption{Reconstruction success in terms of MSE across the datasets using different privacy measures. The left hand figure shows that, for the same epsilon, different mechanisms have different MSE values (a measure of reconstruction success). The right-hand figure shows that, for the same Bayes' capacity, different mechanisms have similar MSE values. Thus, the Bayes' capacity aligns better with reconstruction success when comparing mechanisms.}\label{fig:results}
    \end{figure*}

We next examine an alternative mechanism for DP-SGD: the von Mises-Fisher mechanism (VMF). This mechanism was introduced by Weggenmann et al.~\cite{weggenmann-kerschbaum:2021:CCS} and can be applied to high-dimensional vectors much like the Gaussian. (Further details are provided in \App{app:vmf}.) Our goal is to identify whether the VMF mechanism may provide better protection than the Gaussian against reconstruction attacks in FL.

% Although the VMF mechanism satisfies $d$-privacy~\footnote{$d$-privacy~\cite{chatzikokolakis-etal:2013:PETS} is a generalisation of standard differential privacy, most well-known for its application to geo-indistinguishability~\cite{andres2013geo}.} for the angular distance metric $d_\theta$, we provide a conversion to $(\epsilon, \delta)$-DP for comparison with the Gaussian. 
\Alg{alg:sgd2} represents a DP-SGD algorithm designed for the VMF mechanism. We note that the VMF mechanism requires scaling of the final vector to ensure that it resides on the unit sphere $\mathbb{S}^{p-1}$.~\footnote{Here $p$ is computed as the number of weights in the network.} As for the Gaussian mechanism, we find that the DP-SGD algorithm for VMF decomposes into the same deterministic channel $C$ composed with a channel $D$ described by lines 10-11 of \Alg{alg:sgd2}. We now compute the Bayes' capacity for the channel $D$ using the VMF mechanism.

%
%Note that it differs from \Alg{alg:sgd} in that the gradients are scaled rather than clipped; this is due to the design of the VMF mechanism which expects input vectors to sit on the surface of a unit sphere. This deterministic step does not affect the Bayes' capacity (\Lem{lem:det}), so in the following we compute the Bayes' capacity for  line 8 of \Alg{alg:sgd2}.

\begin{theorem}[Bayes' Capacity for VMF]
Let $V_{p, \kappa}:\mathcal{X} \rightarrow \Dist{\mathbb{S}^{p-1}}$ be the mechanism which takes as input a $p$-dimensional vector $x \in \mathcal{X}$ and outputs a perturbed vector $y \in \mathbb{S}^{p-1}$ by applying VMF noise with parameter $\kappa$ on the unit sphere $\mathbb{S}^{p-1}$. Then the Bayes' capacity of $V_{p, \kappa}$ is given by
\[
    \CBayes(V_{p, \kappa}) ~=~ 2 \Gamma^{-1}\left(\frac{p}{2}\right) \frac{\kappa^{\frac{p}{2} - 1}}{2^{\frac{p}{2}} I_{\left(\frac{p}{2} - 1\right)}(\kappa)} e^{\kappa}
\]
where $I_\nu$ is the modified Bessel function of the first kind of order $\nu$.
%\begin{proof}
%See \App{app:proofs}
%\end{proof}
\end{theorem}

\subsection{Experimental results}

We performed experiments using both versions of DP-SGD against the Inverting Gradients Attack\footnote{https://github.com/JonasGeiping/invertinggradients} which is a well-known reconstruction attack against federated learning. For data we used standard training and test images from the \textbf{MNIST} \cite{deng2012mnist} and  \textbf{Fashion-MNIST} \cite{Xiao2017FashionMNISTAN} datasets. For ease of testing used a simplified neural network model consisting of 13,700 parameters. We varied the $\epsilon$ parameters of both mechanisms (for fixed $\delta$)~\footnote{Both the Gaussian and the VMF mechanisms use comparable $\epsilon, \delta$ guarantees, although the proof of this is beyond the scope of this paper.} and measured the attack success using the mean-squared error (MSE), widely used as a measure for reconstruction success~\cite{sun-etal:2023:NeurIPS}. Results are shown in \Fig{fig:results}.

The plot for Epsilon vs MSE shows that the protection provided by the Gaussian mechanism decreases as $\epsilon$ increases (small MSE means more attack success), while the MSE of the VMF remains relatively stable. However, the results for Bayes' capacity vs MSE show that smaller values for the Bayes' capacity (which correspond to less leakage and thus better protection) also correspond to better effectiveness against reconstruction attacks (larger MSE), regardless of the mechanism used. Thus, Bayes' capacity is a better proxy for reconstruction attack effectiveness than the $\epsilon$ of DP.

\begin{figure}[th]
    \centering
    \begin{tabular}{ccc}                     
        \subfloat[Original]{\includegraphics[width=0.08\textwidth]{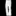}}  
        &         
        \subfloat[Gaussian ]{\includegraphics[width=0.08\textwidth]{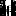}}  
        &         
        \subfloat[VMF]{\includegraphics[width=0.08\textwidth]{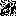}}   \\
          \subfloat[Original]{\includegraphics[width=0.08\textwidth]{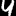}}  
        &         
        \subfloat[Gaussian ]{\includegraphics[width=0.08\textwidth]{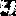}}  
        &         
        \subfloat[VMF]{\includegraphics[width=0.08\textwidth]{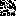}}
\end{tabular}
   % \vspace{4mm}
    \caption{Example of reconstruction success for FMNIST (top) and MNIST (bottom) datasets using $\epsilon = 173$ for the Gaussian and VMF mechanisms.}\label{fig:example}
    \end{figure}

\Fig{fig:example} illustrates the success of the reconstruction attack when using the Gaussian vs the VMF mechanism for $\epsilon = 173$ on both mechanisms. This shows that the Gaussian provides less protection than the VMF mechanism for the same $\epsilon$.

% Importantly, when comparing these mechanisms, the $\epsilon$ value is not indicative of the protections provided against the reconstruction attack.
%This suggests that the Bayes' capacity can be used as a more reliable method of comparing mechanisms with respect to their reconstruction attack protections.

\section{Discussion and Conclusion}

We have shown that Bayes' capacity measures the leakage of an FL system incorporating DP-SGD to a Bayesian attacker whose goal is to perform a perfect reconstruction attack. We have provided a continuous version of Bayes' capacity and derived its formulation for 2 ``incomparable'' differential privacy mechanisms. Experimentally we have demonstrated that the Bayes' capacity provides a better measure for reconstruction attack success than does the $\epsilon$ of differential privacy, even when ``approximate'' measures for a successful attack are used.

The lack of a well-defined reconstruction success measure for image reconstruction makes it more difficult to identify a clear leakage measure corresponding to a reconstruction attack. Indeed, although Bayes' capacity is a robust measure for leakage, and can be used for comparison of mechanisms wrt.\ an exact reconstruction attack, we have questions regarding its robustness for comparing mechanisms wrt.\ other measures. For example, if $\CBayes(C) < \CBayes(D)$ then it does not necessarily follow that $\mathcal{L}_{\textit{MSE}}(\upsilon, C) < \mathcal{L}_{\textit{MSE}}(\upsilon, D)$, even though in our experiments we found this to be the case. We leave further investigation of this to future work.

%From our empirical evaluation, our main conclusions are that the Gaussian mechanism provides consistently higher utility but the VMF provides better privacy and defence against reconstruction attacks. However, when we empirically match the utilities for VMF and Gaussian mechanisms, the protection provided by VMF is greater, implying that the VMF mechanism offers a better privacy-utility trade-off than the Gaussian. Furthermore, we found experimentally that Bayes' capacity does in fact correlate with the reconstruction effectiveness independently of the mechanism, unlike the epsilon values in differential privacy which are mechanism-dependent. This suggests that indeed Bayes' capacity is a better predictor of reconstruction attack success than is the epsilon of differential privacy.

%indicate that epsilon as a measure of privacy is not a good predictor for reconstruction accuracy. Indeed, the results for MSE 

\bibliographystyle{IEEEtran}
\bibliography{IEEEabrv,references}

\appendices

\section{The VMF Mechanism}\label{app:vmf}

Weggenmann et al.~\cite{weggenmann-kerschbaum:2021:CCS} introduced the VMF mechanism derived from the VMF distribution that perturbs an input vector $x$ on $\mathbb{S}^{p-1}$. They show that this mechanism satisfies $\epsilon d_\theta$-privacy, where $d_\theta$ is the angular distance between vectors, and $\epsilon$ is the scaling parameter corresponding to the $\kappa$ parameter in the definition of the VMF. %the unit sphere in a $p$-dimensional space, denoted by $\mathbb{S}^{p-1}$. That is, $x\in \mathbb{S}^{p-1}$ if and only if $\| x\|_2 = 1$.

\begin{definition}[VMF mechanism]
For any secret $x \in \mathbb{S}^{p-1}$ and an output $y\in\mathbb{S}^{p-1}$, the corresponding probability density function of the VMF mechanism centred around $x$ with a concentration parameter $\kappa > 0$ is given by

\begin{equation}\label{eq:VMF}
    f_{p,\kappa}^V(x)(y)~ = ~ \frac{1}{C_{p,\kappa}}e^{\kappa x^T y}~,
\end{equation}
where superscript $T$ denotes vector transpose, $C_{p,\kappa} = \frac{(2\pi)^{\nu+1}I_\nu(\kappa)}{\kappa^\nu}$ is the normalisation factor, $\nu = p/2-1$, and $I_\nu$ is the modified Bessel function of the first kind of order $\nu$. 
\end{definition}

The concentration parameter $\kappa$ roughly corresponds to the variance of the VMF around its mean; larger $\kappa$ corresponds to a smaller variance.

\section{Proofs}\label{app:proofs}

This section provides details of proofs for theorems presented in the paper.

\begin{lemma}
Let $C$, $D$ be channels such that $C{\cdot}D$ is defined, and $C$ is deterministic. Then $\CBayes(C{\cdot}D) = \CBayes(D)$.
\begin{proof}
Preprocessing by a deterministic channel has the effect of swapping rows or repeating rows. Neither of these actions involves removing a row of a channel or modifying the maximum element in the column of a channel. (Note that when we define channels we exclude the possibility of all zero columns - thus $C$ cannot remove rows from $D$). Since $\CBayes$ sums the column maxes for each column, it is unaffected by the preprocessing by $C$. The result follows.
\end{proof}
\end{lemma}

\begin{theorem}[Bayes' Capacity for Gaussian]
   Let $G_{p, \sigma}: \mathcal{X} \rightarrow \Dist{\mathbb{R}^{p}}$ be the mechanism which takes as input a $p$-dimensional vector $x \in \mathcal{X}$ and outputs a perturbed vector $y \in \mathbb{R}^{p}$ by applying Gaussian noise with parameter $\sigma$ to each element of the (clipped) input vector with clipping length $c=1$, and then averaging by $L$. Then the Bayes' capacity of $G_{p, \sigma}$ is given by:
\[
    \CBayes(G_{p, \sigma}) ~=~ \frac{2}{\Gamma\left(\frac{p}{2}\right) 2^{\frac{p}{2}} \sigma^p} Z + \frac{R^p}{\Gamma\left(\frac{p}{2} + 1\right) 2^{\frac{p}{2}} \sigma^{p}}
\]
where $R$ is the number of layers in the network (each clipped to length 1) and $Z = \sum_{i=0}^{p-1} \Gamma(\frac{p-i}{2}) (\sqrt{2} \sigma)^{p-i} { p-1 \choose i } R^i$.

\begin{proof}{(Sketch)}
  We are given a domain of $p$-dimensional vectors $\mathcal{X}$ which are mapped by \Alg{alg:sgd} to output vectors in $\mathbb{R}^p$ after clipping. We remark that the effect of clipping in \Alg{alg:sgd} is that, given an unclipped $x \in \mathcal{X}$ and its clipped version $x_C \in \mathbb{B}_R^p$, the probability density function $f(x_C)$ centred at $x_C$ is exactly $f(x)$ for noise-adding mechanism $f$. Therefore we can write: 
\begin{align*}
    \CBayes(G_{p, \sigma}) ~&= \int_{\mathbb{R}^p} \sup_x G_{p, \sigma}(x)(y) ~dy \\
    &= \int_{y \in \mathbb{B}_R^{p}} \sup_x G_{p, \sigma}(x)(y) ~dy ~+~ \\
    & ~~~~~~~~~ \int_{y \in \mathbb{R}^p \setminus \mathbb{B}_R^{p}} \sup_x G_{p, \sigma}(x)(y) ~dy
\end{align*}
For the first integral, the supremum for each $y$ occurs when $y = x$. Using the notation $f_{p,\sigma^2}^G(x)$ to describe the probability density function for the Gaussian mechanism, we have:
\begin{align*}
    \int_{y \in \mathbb{B}_R^{p}} \sup_x G_{p, \sigma}(x)(y) ~dy &= \int_{y \in \mathbb{B}_R^{p}} f_{p,\sigma^2}^G(y)(y) ~dy \\
    &= f_{p,\sigma^2}^G(u)(u) \int_{y \in \mathbb{B}_R^{p}} ~dy \\
    &= \frac{1}{\sqrt{2\pi\sigma^2}^p} V(\mathbb{B}_R^p)
\end{align*}
where $u$ is an arbitrary vector inside $\mathbb{B}_R^p$ and $V(S)$ denotes the volume of $S$.

For the second integral, we observe that the supremum at $y$ occurs at the point $x$ on the surface of $\mathbb{B}_R^{p}$ which minimises the distance between $x$ and $y$. 
This occurs when $x$ is on the ray from $y$ through the origin. i.e. $x = R \frac{y}{\|y\|_2}$. Thus we can write: 

\begin{align*}
    \int_{y \in \mathbb{R}^p \setminus \mathbb{B}_R^{p}} &\sup_x G_{p, \sigma}(x)(y) ~dy = \int_{y \in \mathbb{R}^p \setminus \mathbb{B}_R^{p}} f_{p,\sigma^2}^G\left(R \frac{y}{\|y\|_2}\right)(y) dy \\
    &= \int_{\mathbb{R}^p \setminus \mathbb{B}_R^{p}} \frac{1}{\sqrt{2\pi\sigma^2}^p} e^{\frac{-\|y - R \frac{y}{\|y\|_2}\|_2^2}{2 \sigma^2}} dy \\
    &= \int_{\mathbb{R}^p \setminus \mathbb{B}_R^{p}} \frac{1}{\sqrt{2\pi\sigma^2}^p} e^{\frac{-(\|y\|_2 - R)^2}{2 \sigma^2}} dy\\
    &= \frac{1}{\sqrt{2\pi\sigma^2}^p} A(\mathbb{S}^{p-1}) \int_{r=R}^{\infty} r^{p-1} e^{\frac{-(r-R)^2}{2 \sigma^2}} dr \\
    &= \frac{1}{\sqrt{2\pi\sigma^2}^p} A(\mathbb{S}^{p-1}) \int_{s=0}^{\infty} (s+R)^{p-1} e^{\frac{-s^2}{2 \sigma^2}} ds 
\end{align*}
where $A(S)$ denotes the area of $(S)$ and $R$ is the radius of the ball $\mathbb{B}_R^p$. Note that the second-last line follows from change of variables to spherical coordinates.

Now, expanding $(s+R)^{p-1}$ and using the identity $\int_0^\infty x^{k-1} e^{\frac{-x^2}{2 \sigma^2}} dx = \frac{\Gamma(\frac{k}{2})}{2} (2 \sigma^2) ^ {\frac{k}{2}}$, we compute: 
\begin{align*}
     \int_{s=0}^{\infty} (s+R)^{p-1} e^{\frac{-s^2}{2 \sigma^2}} ds 
    &= \frac{1}{2} \sum_{i=0}^{p-1} \Gamma\left(\frac{p-i}{2}\right) (\sqrt{2} \sigma)^{p-i} { p-1 \choose i } R^i
\end{align*}
The result follows.
\end{proof}
\end{theorem}

\begin{theorem}[Bayes' Capacity for VMF]
Let $V_{p, \kappa}:\mathcal{X} \rightarrow \Dist{\mathbb{S}^{p-1}}$ be the mechanism which takes as input a $p$-dimensional vector $x \in \mathcal{X}$ and outputs a perturbed vector $y \in \mathbb{S}^{p-1}$ by applying VMF noise with parameter $\kappa$ on the unit sphere $\mathbb{S}^{p-1}$. Then the Bayes' capacity of $V_{p, \kappa}$ is given by
\[
    \CBayes(V_{p, \kappa}) ~=~ 2 \Gamma^{-1}\left(\frac{p}{2}\right) \frac{\kappa^{\frac{p}{2} - 1}}{2^{\frac{p}{2}} I_{\left(\frac{p}{2} - 1\right)}(\kappa)} e^{\kappa}
\]
where $I_\nu$ is the modified Bessel function of the first kind of order $\nu$.
\begin{proof}{(Sketch)}
In \Alg{alg:sgd2} the domain $\mathcal{X}$ of gradient vectors is $p$-dimensional vectors in $\mathbb{R}^p$ which are subsequently scaled to unit length so that the output domain is restricted to vectors in $\mathbb{S}^{p-1}$. For each input $x \in \mathcal{X}$ the VMF noise mechanism attains its maximum value at the point $y = x / \|x\|_2$. Since $\mathbb{S}^{p-1} \subseteq \mathcal{X}$, from \Eqn{eqn:bcapacity} we can choose each pointwise supremum over $\mathcal{X}$ at $y = x$. Using the notation $f_{p,\kappa}^V(x)$ from \Eqn{eq:VMF} describing the VMF density function, we have:
\begin{align*}
    \CBayes(V_{p, \kappa}) ~&= \int_{\mathbb{S}^{p-1}} \sup_x V_{p, \kappa}(x)(y) ~dy \\
    &= \int_{y \in \mathbb{S}^{p-1}} f_{p,\kappa}^V(y)(y)~ dy \\
    &= f_{p,\kappa}^V(u)(u) \int_{y \in \mathbb{S}^{p-1}} ~dy \\
    &= \frac{1}{C_{p,\kappa}}e^{\kappa} A(\mathbb{S}^{p-1})
\end{align*}
where $u$ denotes an arbitrary point on $\mathbb{S}^{p-1}$ (noting that the value of the VMF at any $u$ is independent of the mean $u$ chosen) and $A(S)$ denotes the area of $S$.
The result follows.
\end{proof}
\end{theorem}

\end{document}